\documentclass[runningheads]{llncs}
\usepackage{graphicx}
\usepackage{comment}
\usepackage{amsmath,amssymb} 
\usepackage{color}

\usepackage[width=122mm,left=12mm,paperwidth=146mm,height=193mm,top=12mm,paperheight=217mm]{geometry}

\usepackage{xspace}
\makeatletter

\DeclareRobustCommand\onedot{\futurelet\@let@token\@onedot}
\def\@onedot{\ifx\@let@token.\else.\null\fi\xspace}

\def\eg{\emph{e.g}\onedot} 
\def\ie{\emph{i.e}\onedot} 
 
\def\etc{\emph{etc}\onedot} 
 
\def\etal{\emph{et al}\onedot}
\makeatother

\usepackage{algorithm}
\usepackage{algorithmic}

\begin{document}
\pagestyle{headings}
\mainmatter
\def\ECCVSubNumber{13}  

\title{Optical Flow Distillation: Towards Efficient and Stable Video Style Transfer} 

\titlerunning{Optical Flow Distillation: Towards Efficient and Stable Video Style Transfer}
%
\author{Xinghao Chen\inst{1\star} \and
Yiman Zhang\inst{1\star} \and
Yunhe Wang\inst{1} \and Han Shu\inst{1} \and \\ Chunjing Xu\inst{1} \and Chang Xu\inst{2}}
%
\authorrunning{X. Chen et al.}
%
\institute{Noah's Ark Lab, Huawei Technologies \and
School of Computer Science, Faculty of Engineering, University of Sydney
\email{\{xinghao.chen,zhangyiman1,yunhe.wang,han.shu,xuchunjing\}@huawei.com},
\email{c.xu@sydney.edu.au}}
\maketitle

\renewcommand{\thefootnote}{\fnsymbol{footnote}}
\footnotetext[1]{Equal Contribution.}

\begin{abstract}
Video style transfer techniques inspire many exciting applications on mobile devices. However, their efficiency and stability are still far from satisfactory. To boost the transfer stability across frames, optical flow is widely adopted, despite its high computational complexity, \eg occupying over 97\% inference time. This paper proposes to learn a lightweight video style transfer network via knowledge distillation paradigm. We adopt two teacher networks, one of which takes optical flow during inference while the other does not. The output difference between these two teacher networks highlights the improvements made by optical flow, which is then adopted to distill the target student network. Furthermore, a low-rank distillation loss is employed to stabilize the output of student network by mimicking the rank of input videos. Extensive experiments demonstrate that our student network without an optical flow module is still able to generate stable video and runs much faster than the teacher network.
\end{abstract}

\section{Introduction}

Artistic style transfer aims to transform the artistic style of a given painting to an image and has attracted tremendous interests since the seminal work of Gatys~\etal~\cite{gatys2015neural}. Plenty of works have been dedicated to improving the performance of single image style transfer from different perspectives~\cite{lu2017decoder,johnson2016perceptual,li2017universal,Lu_2019_ICCV,liao2017visual,chen2017stylebank,dumoulin2017learned}. Meanwhile, there is growing attention for video style transfer~\cite{huang2017real,gao2018reconet,ruder2016artistic,ruder2018artistic,gao2020fast,wang2020consistent} due to its wider application scenarios, \eg,  movie synthesis and mobile applications. Compared with single image style transfer, stylizing a video is a much more challenging task. The key problem is the flickering phenomenon of the stylized videos. Due to the motion of objects and the changing of light in the video \etc, transferring the videos frame-by-frame independently causes the temporal inconsistency between consecutive stylized frames.

\begin{figure*}[t]
	\centering
	\includegraphics[width = 0.9\linewidth]{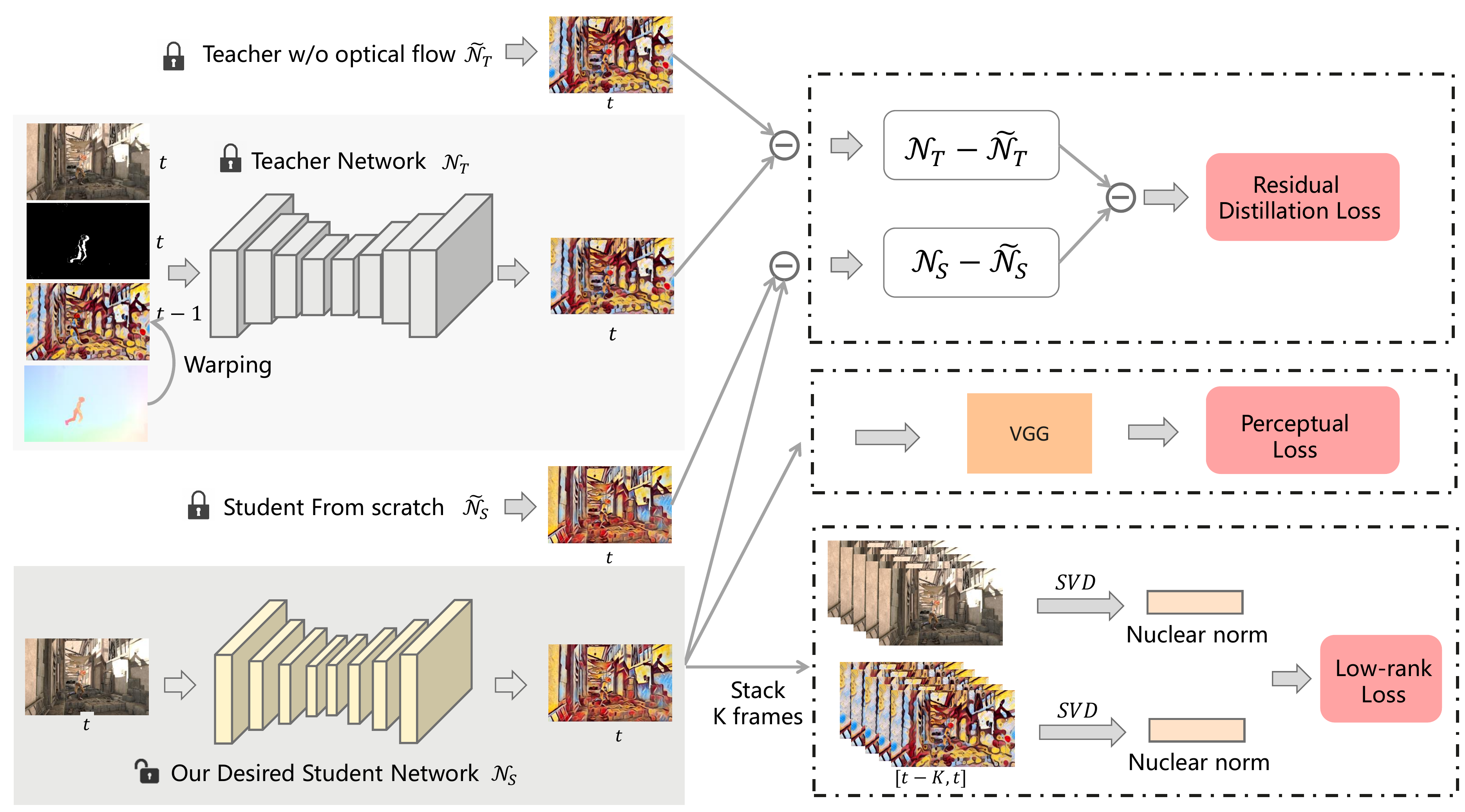}
	\caption{The diagram of our proposed method. Our optical flow distillation method transfers knowledge from a stable video style transfer network $\mathcal{N}_T$ with optical flow to a lightweight network $\mathcal{N}_S$ that does not require optical flow during inference. The \textit{residual distillation loss} encourages the student network to learn knowledge of the difference between the teacher networks with and without optical flow ($\mathcal{N}_T - \tilde{\mathcal{N}}_T$). Additionally, the \textit{low rank distillation loss} is exploited to stabilize the output stylized videos of student network by mimicking the low rank of input videos. The basic \textit{perceptual loss} is also used for style transfer.}
	\label{fig:diagram}
\end{figure*}
To tackle the problem discussed above, Ruder~\etal~\cite{ruder2016artistic,ruder2018artistic} proposed a network that takes several inputs, including the current frame, occlusion mask and the previous stylized image warped by the optical flow. Despite producing smooth and coherent stylized videos, the inference speeds of these methods are relatively slow due to the calculation of optical flow on the fly. For example, the method of Ruder~\etal~\cite{ruder2018artistic} takes about $210 ms$ to process a frame of $640 \times 320$ in the video, among which almost $97\%$ of time is related to the time-consuming calculation of optical flow and warping operation. 

Many methods have been proposed to alleviate the burden of on-the-fly optical flow computation and have achieved real-time speed for video style transfer~\cite{huang2017real,gupta2017characterizing,gao2018reconet}. These methods utilize temporal consistent loss guided by pre-computed optical flow in training, which encourages the model to learn smooth and coherent stylized images in consecutive video frames. These methods are faster since they only take the current frame as input and get rid of optical flow estimation in the inference stage. Despite the fast speed, they still have less stable results when compared with methods that adopt optical flow during inference phase, as also indicated in some prior literature~\cite{gao2018reconet,huang2017real}. In addition, there are a number of model compression and speeding-up methods, \eg,  pruning~\cite{he2017channel,rethinking-pruning,shu2019co}, quantization~\cite{zhou2016dorefa,birealnet,shen2019searching}, distillation~\cite{romero2015fitnets,zagoruyko2017paying,liu2019structured} and neural architecture search~\cite{gong2019autogan,yang2020cars}. Most of existing methods are explored for single image processing or recognition models. An efficient algorithm for learning efficient and stable video style transfer networks is urgently required.

In this paper, we present a novel knowledge distillation method to achieve a better trade-off between inference speed and stability of the stylized videos. The framework of our proposed method is depicted in Fig.~\ref{fig:diagram}. We choose a stable video style transfer network including an optical flow module in the inference phase as the teacher network, and a lightweight network only consumes current frame as the desired student network. 
We propose the residual distillation loss to encourage the student network to learn the residual between output stylized videos produced by teacher network with and without optical flow in inference. Moreover, motivated by the fact that consistent frames have the properties of low rank, we add additional low rank loss so that student network produces coherent stylized videos that have similar low rank with input videos.
The inference speed of the student network is significantly faster than that of the teacher network after removing the optical flow module.

We then carefully design the evaluation experiments on benchmarks. The results illustrate that videos generated by student network learned using the proposed optical flow distillation paradigm have similar visualization quality to that of the teacher model but obviously lower computational costs, which can be further launched in real-time on mobile devices.

The rest of the paper is organized as follows: Section~\ref{sec:related} investigates the state-of-the-art neural style transfer methods and knowledge distillation approaches. Section~\ref{sec:distill} elaborates the proposed knowledge distillation for real-time video style transfer. In Section~\ref{sec:exp}, we provide extensive experiments to compare with state-of-the-art methods and perform ablation studies. Section~\ref{sec:con} provides a brief conclusion of this paper and discusses future work.

\section{Related Work}
\label{sec:related}
In this section, we briefly review the related work about neural style transfer and knowledge distillation. 

\subsection{Video Style Transfer}
Neural style transfer is one of the  most popular research hotspots in recent years. Gatys~\textit{et al.}~\cite{gatys2015neural,gatys2016image} used CNN to iteratively reconstruct a stylized image by minimizing the difference between the target image, the content image and the style image in high-level features. These methods solve the optimization by backward propagation and are computationally expensive. To make the inference more efficient, Johnson~\textit{et al.}~\cite{johnson2016perceptual} proposed a feed-forward network to stylize images, which replaces the iterative process of optimizing pictures with the optimization of CNNs via training. 

Video style transfer is attracting more and more research interests. Researchers tried to utilize the inter-frame temporal relation to improve the visual stability of stylized videos, specifically motion estimation based on optical flow.
Ruder~\textit{et al.}~\cite{ruder2016artistic} initialized the optimization of the current frame with stylized output of the previous frame and proposed temporal loss which uses optical flow to maintain inter-frame consistency. This image based optimization algorithm outputs a very stable video but costs about 3 minutes to process a frame even with precomputed optical flow. Therefore, fast video style transfer is mainly based on model optimization.
Ruder~\textit{et al.}~\cite{ruder2018artistic} proposed a framework to use optical flow both in the training stage and in the inference stage to improve temporal consistency of output stylized videos. This framework contains two networks.
The first network obtains the first frame of the video as input and outputs the stylized result.
The second network obtains three inputs, including the current frame, the previous stylized frame warped by the optical flow and the mask which indicates
motion boundaries and outputs the stylized result of current frame. 
Similarly, the architecture in \cite{chen2017coherent} 
utilized optical flow both in training stage and inference stage.  All these methods~\cite{ruder2018artistic,chen2017coherent} got stable stylized video but can not be used for real-time video style transfer.  
To address this problem, another family of video style transfer methods~\cite{gao2018reconet,huang2017real} utilize optical flow only in the training stage thus speed up the inference. These methods utilize similar temporal loss to train the feed-forward transform network to improve the temporal consistency of output videos. They get rid of computing optical flows on the fly but produce less stable stylized results than those networks that adopt optical flows during inference stage~\cite{ruder2018artistic,huang2017real}. Our method aims to mitigate the gap between optical flow based and optical flow free methods for video style transfer via knowledge distillation.

\subsection{Knowledge Distillation}
Knowledge distillation is a technique that leverages intrinsic information of teacher network to train a smaller one, which is first pioneered by Hinton~\etal~\cite{hinton2015distilling}. Since then many algorithms have been proposed to improve knowledge distillation~\cite{huang2017like,yim2017gift,chen2019data,romero2015fitnets,heo2019comprehensive}. Wang~\etal~\cite{wang2018adversarial} exploited generative adversarial network to encourage the student network to learn similar feature distribution with teacher network. 
Zagoruyko and Komodakis~\cite{zagoruyko2017paying} proposed to utilize spatial attention for distilling intermediate latent features of the network. Heo~\textit{et al.}~\cite{heo2019knowledge} proposed a novel activation transfer loss to distill knowledge of activation boundaries from the teacher network.
Chen~\etal~\cite{chen2018learning} introduced the locality preserving loss to preserve relationships between samples in high dimensional features from teacher network and low dimensional features from student network. 
Knowledge distillation has also been adopted in many other applications, \eg,  object detection~\cite{li2017mimicking,wei2018quantization,chen2017learning}, semantic segmentation~\cite{jiao2019geometry,liu2019structured,he2019knowledge} and pose regression~\cite{saputra2019distilling,wang2019distill}. However, distilling knowledge from a stable video style transfer to a lightweight student network is not yet explored, which is the main purpose of this paper.

\section{Method}
\label{sec:distill}

Let $\mathcal{N}_T$ and $\mathcal{N}_S$ denote the teacher network and the desired student network, respectively. The teacher network~\cite{ruder2018artistic} utilizes optical flow in both training and inference phase to increase the stability of video neural style transfer. Since the teacher network needs to calculate optical flow in inference phase, it is time-consuming and not suitable for real-time applications. Thus, the student network is expected to have no optical flow module. The lightweight student network gets rid of optical flow estimation in inference phase and thus runs much faster. We choose similar network architecture with ReCoNet~\cite{gao2018reconet} as our student network. The goal of the proposed optical flow knowledge distillation is to train the student network $\mathcal{N}_S$
with aid of teacher network  $\mathcal{N}_T$, so that $\mathcal{N}_S$ obtains similar stability as $\mathcal{N}_T$ yet still runs fast since $\mathcal{N}_S$ does not compute optical flow on-the-fly during inference.

\subsection{Preliminaries: Style Transfer Loss}
Here we briefly revisit the perceptual loss introduced by Johnson~\etal~\cite{johnson2016perceptual}, which has been widely used for style transfer algorithms. The perceptual loss includes the content loss to encourage the output stylized image to have similar content representations with the input image, and  the style loss to capture the style information. In addition, total variation regularization ($\mathcal{L}_{tv}$) is generally introduced to encourage spatial smoothness in the stylized images. Therefore, the basic style transfer perceptual loss contains three terms:
\begin{equation}\label{eq:percep}
\mathcal{L}_{percep}(x, I_{s}) = \lambda_c \mathcal{L}_{content}(x, I_{s}) + \lambda_s \mathcal{L}_{style}(x, I_{s}) + \lambda_{tv} \mathcal{L}_{tv}(x, I_{s}),
\end{equation}
where $I_s$ is the given style image and $x$ is an input image. $\lambda_c$, $\lambda_s$ and $\lambda_{tv}$ are hyper parameters to balance three different losses. To simplify the notations, we omit the $I_s$ and denote the perceptual loss as $\mathcal{L}_{percep}(x)$ in the following sections.

This basic style transfer loss can be exploited to train a student network from scratch, which is denoted as $\tilde{\mathcal{N}}_S$. Since $\tilde{\mathcal{N}}_S$ is trained frame by frame, it suffers from flickering in consecutive frames and produces unstable stylized videos. Our goal is transferring the knowledge of a teacher network $\mathcal{N}_T$  to train the desired student network $\mathcal{N}_S$, so that student network produces coherent stylized videos.

\subsection{Residual Distillation Loss}
\label{subsec:res_kd}

\begin{figure}[t]
	\centering
	\includegraphics[width = 0.8\linewidth]{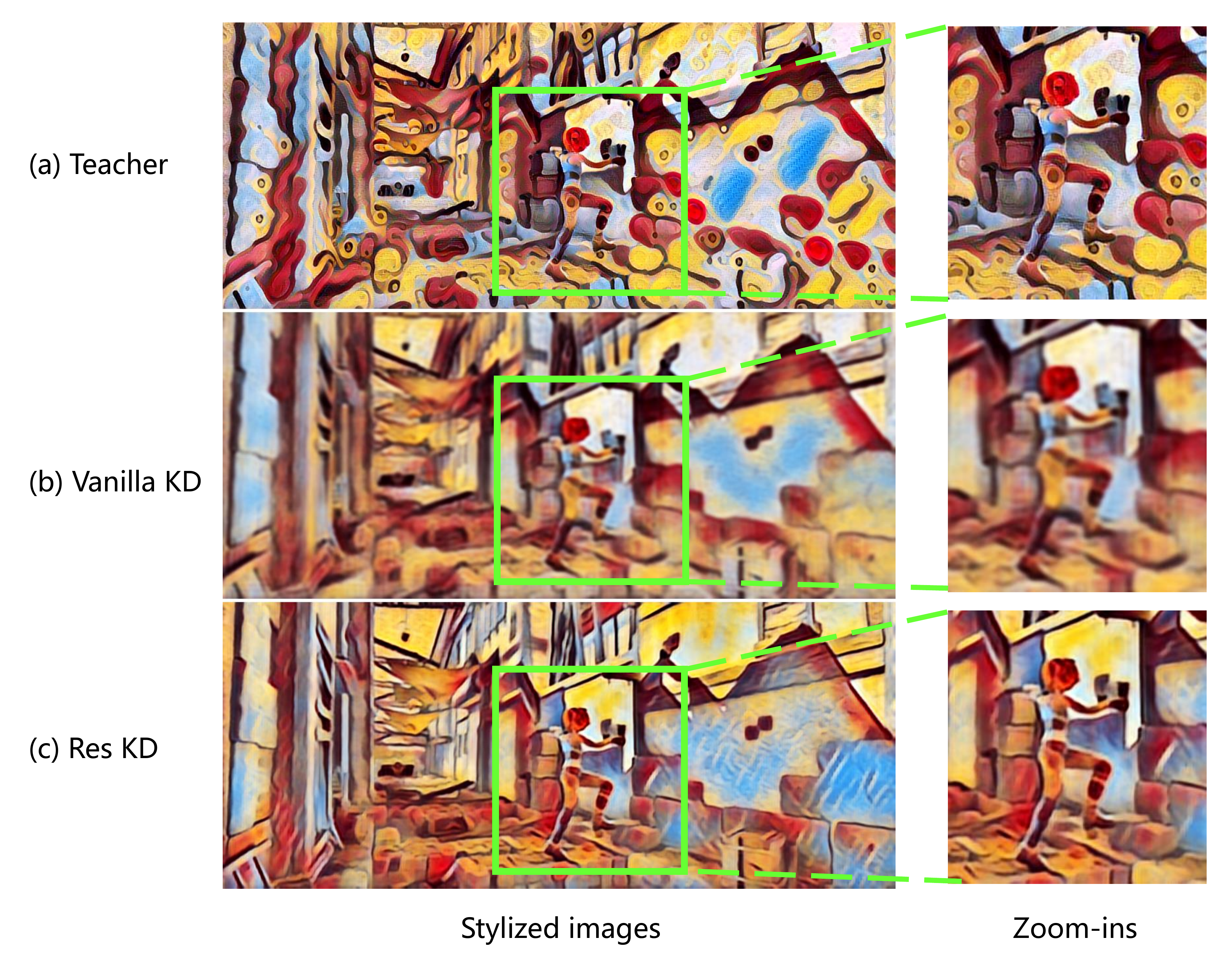}
	\caption{(a) Output stylized frames from teacher network, (b) Results of student network with vanilla distillation loss and (c) Results of student network with residual distillation loss.}
	\label{fig:res_kd}
\end{figure}

A straightforward way to train the desired student network $\mathcal{N}_S$ via knowledge distillation is directly using the output stylized image of teacher network to teach the student network, as shown in the following loss function:
\begin{equation}\label{eq:mse}
\mathcal{L}_{vanilla}(x) = ||\mathcal{N}_T (x, f) - \mathcal{N}_S (x)||^2_2,
\end{equation}
where $\mathcal{N}_T (x, f)$ and $\mathcal{N}_S (x)$ are the output stylized images for input image $x$ of teacher network and student network respectively, $f$ is the corresponding optical flow.

However, this strategy produces blurred stylized images, as shown in Fig.~\ref{fig:res_kd}. The reasons behind are two folds. Firstly, it is a challenging optimization problem to force the output of student network to align with that of the teacher network at pixel level. More importantly, due to the difference of network architectures between student network and teacher network, the style of the output images from student network may slightly differ from those produced by teacher network. Therefore, directly distilling knowledge from the output stylized images of teacher network is hard to train a good student network.

To address the above problem, we propose the residual distillation loss for better knowledge distillation. Our goal is to train an optical flow free student network from the knowledge of an optical flow based teacher network. Therefore, the knowledge of the difference between teacher networks with and without optical flow is the key information to train a stable student network.
Let $\tilde{\mathcal{N}}_T$ denotes the model similar to teacher model $\mathcal{N}_T$ but does not adopt optical flow for video style transfer and $ \tilde{\mathcal{N}}_T (x)$ is the output of $\tilde{\mathcal{N}}_T$ for input image $x$. Obviously the output stylized video flickers since basically $\tilde{\mathcal{N}}_T$ just predicts stylized images frame by frame. The difference between the output of $\mathcal{N}_T$ and $\tilde{\mathcal{N}}_T$ is:
\begin{equation}\label{eq:delta_T}
\Delta \mathcal{T} (x) =  \mathcal{N}_T (x, f) - \tilde{\mathcal{N}}_T (x).
\end{equation}
$\Delta \mathcal{T} (x)$ is attributed to additional temporal consistency that is brought by optical flows to the output stylized videos, which is the key information for student network to improve the stability.

$\tilde{\mathcal{N}_{S}}$ is the student network that is trained only using basic style transfer loss as in Eq.~(\ref{eq:percep}) and $\mathcal{N}_S$ is the stable student network we want to obtain. Thus the improvement of $\mathcal{N}_S$ over a vanilla baseline is:
\begin{equation}\label{eq:delta_S}
\Delta \mathcal{S} (x) =  \mathcal{N}_S (x) - \tilde{\mathcal{N}}_S (x).
\end{equation}
We encourage the student network to learn how to improve the stability of output videos over the baseline model by forcing $ \Delta \mathcal{S}$ to imitate $\Delta \mathcal{T}$. The residual distillation loss is formulated as follow:
\begin{equation}\label{eq:res}
\mathcal{L}_{res}(x) =  ||\Delta \mathcal{T} (x) - \Delta \mathcal{S} (x)||^2_2.
\end{equation}
The benefits of above residual distillation loss are two folds. Firstly, $\Delta \mathcal{T} (x)$ is the key information for the teacher network to become stable and could benefit the temporal consistency of student network. What's more, $\Delta \mathcal{T} (x)$ and $\Delta \mathcal{S} (x)$ have eliminated the difference of stylized images brought by structures, thus it could alleviate the blurring effect as shown in Fig.~\ref{fig:res_kd}.

\subsection{Low-rank Distillation Loss}
\label{subsec:lowrank}
In the above section, Eq.~(\ref{eq:res}) is proposed for distilling knowledge from teacher network for one frame. We further develop a distillation loss by exploiting the temporal property of the video.

For a stylized video that is temporally consistent and stable, the same regions that are not located at the occluded regions or motion boundaries are supposed to have similar stylized patterns and strokes. Therefore, a basic assumption for a stable stylized video is that the non-occluded regions should be low rank representations. 

Suppose we have $K$ consecutive frames $\{x_t\}_{t=1}^K$ and the corresponding optical flows $\{f_t\}_{t=1}^K$, the output stylized frames from the student network are calculated as $\{\mathcal{N}_S (x_t)\}_{t=1}^K$. We warp all stylized frames to a specific frame $\mathcal{N}_S (x_{\tau})$ using the optical flows, where $x_\tau$ is often chosen as the middle frame, \ie,  $\tau=\lfloor K/2 \rfloor$. We denote the warped frames as $\mathcal{W}_{t \rightarrow \tau}({\mathcal{N}}_S (x_{t}), f_t)$, where $\mathcal{W}_{t \rightarrow \tau}(\cdot)$ means warping the $t^{th}$ frame to $\tau ^{th}$ frame. $\mathcal{M}_t$ is the occlusion mask for $x_t$, where $0$ indicates the motion regions or boundaries and $1$ indicates the traceable regions.
In this way, if we put all traceable regions $\mathcal{R}_t = \mathcal{M}_t \odot \mathcal{W}_{t \rightarrow \tau}(\mathcal{N}_S (x_{t}), f_t)$ into a matrix $\mathcal{X}$, then $\mathcal{X}$ is low rank, \ie, 
\begin{equation}\label{eq:rankmat}
\mathcal{X} = \left[vec(\mathcal{R}_0), \ldots , vec(\mathcal{R}_K) \right]^T \in \mathbb{R}^{K \times L},
\end{equation}
where $L=H\times W$ is the number of pixels in the images and $vec(\cdot)$ means transforming a two dimensional image into a one-dimensional vector.

We can get the Singular Value Decomposition (SVD) of $\mathcal{X}$ by:
\begin{equation}\label{eq:svd}
\mathcal{X} := \mathcal{U}\Sigma\mathcal{V}^T,
\end{equation}
where $\mathcal{U} \in \mathbb{R}^{K \times K}$ and $\mathcal{V} \in \mathbb{R}^{L \times L}$ are orthogonal matrices and $\Sigma \in \mathbb{R}^{K \times L}$ is the singular value matrix. The diagonal elements in $\Sigma$ are singular values $\Gamma = \{\gamma_0, \ldots, \gamma_K\}$ of the matrix $\mathcal{X}$. The rank of $\mathcal{X}$ is calculated as:
\begin{equation}\label{eq:rank}
\operatorname{rank}(\mathcal{X}) = \sum_i^K {\mathbb I (\gamma_i > 0)},
\end{equation}
where $\mathbb I (\cdot)$ is the indicator function.

However, the rank of a matrix is a non-differentiable function and can not directly be optimized via CNNs. Therefore, we instead adopt nuclear norm of $\mathcal{X}$, which is a convex relaxation of the rank function. The nuclear norm $||\cdot||_*$ is given by:
\begin{equation}\label{eq:nuclear}
||\mathcal{X}||_* = \sum_i^K \gamma_i.
\end{equation}
where $\gamma_i$ is the $i^{th}$ singular value of $\mathcal{X}$.

We expect that the rank of output stylized videos from student network should be similar with the stable videos. Intuitively, a straightforward way is to encourage the student network to imitate the nuclear norm of teacher network by the following low-rank distillation loss:
\begin{equation}\label{eq:rankdistill2}
\mathcal{L}_{rank}^T = (||\mathcal{X}_T||_* - ||\mathcal{X}_S||_*)^2,
\end{equation}
where $||\mathcal{X}_T||_*$ and $||\mathcal{X}_S||_*$ are nuclear norms of output stylized videos from teacher network and student network, respectively. Distilling knowledge of low rank from teacher network may help to improve the results of student network. However, the stability of output videos from the teacher network is still worse than the input videos. The stability of input video is a better teacher for the student network. To this end, we propose to distill the rank of input videos to train the student network by the following distillation loss:
\begin{equation}\label{eq:rankdistill3}
\mathcal{L}_{rank}^{input} = (||\mathcal{X}_{input}||_* - ||\mathcal{X}_S||_*)^2,
\end{equation}
where $||\mathcal{X}_{input}||_*$ is the nuclear norm of the input video. We utilize $\mathcal{L}_{rank}^{input}$ to train the student network and will further discuss the different low rank losses in experiments.

\subsection{Optimization}
\begin{algorithm}[t]
	\caption{Optical Flow Knowledge Distillation.}
	\label{Alg:main}
	\begin{algorithmic}[1]
		\REQUIRE A given teacher network $\mathcal{N}_T$, a given vanilla teacher network $\tilde{\mathcal{N}}_T$ that is similar with $\mathcal{N}_T$ but does not use optical flow, a student network that is trained from scratch $\tilde{\mathcal{N}}_S$, a style image $I_{style}$, the training set $\{x^i, f^i\}_{i=1}^{N}$ where $x^i = \{x^i_t\}_{t=1}^{K}$ and $f^i = \{f^i_t\}_{t=1}^{K}$ are input images and optical flows, respectively.
		\STATE Initialize a neural network $\mathcal{N}_S$, which does not need to compute optical flows during inference.
		\REPEAT
		\STATE Randomly select a batch of training data $\{x^i, f^i\}_{i=1}^{m}$.
		\STATE \textbf{Residual Distillation Loss}
		\STATE Employ teacher network $\mathcal{N}_T$ and $\tilde{\mathcal{N}}_T$ on the mini-batch and calculate Eq.~(\ref{eq:delta_T}).\\\quad\quad\quad\quad\quad\quad $\Delta T (x)\leftarrow \mathcal{N}_T (x, f) - \tilde{\mathcal{N}}_T (x)$
		\STATE Employ student network $\mathcal{N}_S$ and $\tilde{\mathcal{N}}_S$ on the mini-batch and calculate Eq.~(\ref{eq:delta_S}).\\\quad\quad\quad\quad\quad\quad $\Delta S (x)\leftarrow \mathcal{N}_S (x) - \tilde{\mathcal{N}}_S (x)$
		\STATE Calculate the loss function $\mathcal{L}_{res}$ according to Eq.~(\ref{eq:res}).
		\STATE \textbf{Baseline Loss}
		\STATE Calculate perceptual loss $\mathcal{L}_{percrp}$ according to Eq.~(\ref{eq:percep}).
		\STATE Calculate the temporal loss function $\mathcal{L}_{temp}$ according to Eq.~(\ref{eq:temp_loss}).
		\STATE \textbf{Low-Rank Distillation Loss}
		\STATE Calculate the nuclear norm $||\mathcal{X}_{input}||_*$ of input videos.
		\STATE Calculate the nuclear norm $||\mathcal{X}_{S}||_*$ of output videos from student network $\mathcal{N}_S$.
		\STATE Calculate low rank loss $\mathcal{L}_{rank}^{input}$ according to Eq.~(\ref{eq:rankdistill3}).
		\STATE \textbf{Total Loss and Back Propagation}
		\STATE Calculate the total loss function $\mathcal{L}_{total}$ according to Eq.~(\ref{eq:totalloss}).
		\STATE Update weights in $\mathcal{N}_S$ using gradient descent;
		\UNTIL convergence
		\ENSURE The student network $\mathcal{N}_S$.
	\end{algorithmic}
\end{algorithm}

Temporal consistency loss imposes constraints on consecutive output frames and is widely used in prior methods~\cite{huang2017real,chen2017coherent,gao2018reconet}, which is formulated as follows:
\begin{equation}\label{eq:temp_loss}
\mathcal{L}_{temp}(x_t, x_{t-1}, f_t) = ||\mathcal{M}_t \odot (\mathcal{N}_S(x_t) - \mathcal{W}_{t-1 \rightarrow t}(\mathcal{N}_S(x_{t-1}), f_t) )||^2_2.
\end{equation}
Using temporal loss improves the temporal consistency of student network and thus serves as a stronger baseline than vanilla perceptual loss. We will demonstrate the effectiveness of our proposed method on both two baselines in experiments.

To train student network with perceptual loss or residual distillation loss, the network only needs the current input frame. However, calculating low rank loss and temporal loss needs $K$ consecutive frames. Suppose the input video segments are $x = \{x_t\}_{t=1}^{K}$ and the corresponding optical flows are $f = \{f_t\}_{t=1}^{K}$.
The desired student network can be optimized using the total distillation loss as follow:
\begin{align}\label{eq:totalloss}
\mathcal{L}_{total}(x, f) = &\sum_{t=1}^K (\mathcal{L}_{percep}(x_t) + \lambda_{res} \mathcal{L}_{res}(x_t)) \nonumber\\
&+ \sum_{t=2}^K \lambda_{temp} \mathcal{L}_{temp}(x_t, x_{t-1}, f_t) + \lambda_{rank} \mathcal{L}_{rank}^{input}(x, f), 
\end{align}
where $\lambda_{res}$, $\lambda_{temp}$ and $\lambda_{rank}$ are hyper parameters to balance different terms of losses. 

The $\mathcal{L}_{percep}$ and $\mathcal{L}_{temp}$ in Eq.~(\ref{eq:totalloss}) are used as our baseline for training the student network from scratch. We will explore the baseline with and without the temporal loss. The other two terms are knowledge distillation losses proposed in this paper and we will demonstrate in experiments that these distillation losses improve the stability of output stylized videos.

\section{Experiments}
\label{sec:exp}

In this section, we will demonstrate the effectiveness of our proposed knowledge distillation for lightweight video style transfer network. In addition, we will provide extensive ablation experiments to discuss the impact of different components in our proposed method.

\subsection{Experimental Settings}
We use the Hollywood2 video scene dataset~\cite{marszalek2009actions} as the training data and evaluate our method on MPI Sintel dataset~\cite{butler2012naturalistic}. 
We follow the same data preprocessing methods as in~\cite{ruder2018artistic} and randomly sample 2000 tuples consisting of 5 consecutive frames from Hollywood2 dataset.
MPI Sintel dataset provides ground truth optical flow and occlusion masks, which is widely used for the task of optical flow estimation and is also adopted to evaluate the temporal consistency of video style transfer. Following prior work~\cite{ruder2018artistic,gao2018reconet,huang2017real}, we evaluate our method on five videos in the MPI Sintel dataset.

During training all frames are downscaled to $640 \times 360$ and the input size for evaluation is $1024 \times 436$. We train the student network with learning rate of $10^{-3}$ and a batch size of 1 using Adam optimizer for 10 epochs. The learning rate is decayed by 1.2 in every 500 iterations. The hyper-parameters in Eq.~(\ref{eq:totalloss}) are set to be $K=5$, $\lambda_{res} = 4\times 10^8$, $\lambda_{temp} = 1\times 10^6$ and $\lambda_{rank} = 1\times 10^2$. 

Following the quantitative evaluation metric in prior methods~\cite{gao2018reconet,huang2017real,chen2017coherent}, we utilize $e_{stab}$ to evaluate the temporal consistency of the output stylized videos. $e_{stab}$ is the square root of temporal errors between consecutive frames for the traceable regions of the videos:
\begin{equation}\label{eq:estab}
e_{stab} = \sqrt{\frac 1 N \sum_{t=1}^N \frac 1 D ||\mathcal{M}_t \odot (y_t - \mathcal{W}(y_{t-1}))||_2},
\end{equation}
where $N$ is the number of frames in the testing video, $y_t$ and $y_{t-1}$ are output stylized images for frame $t$ and $t-1$ respectively, $D$ is the number of pixels of output stylized image.

\subsection{Experimental Resutls}

\begin{table}[tb]
	\centering
	\caption{Comparisons of different methods for temporal error $e_{stab}$ and speed (\textit{FPS}) with style \textit{Candy} on five scenes from \textit{MPI Sintel} Dataset. \textsuperscript\textdagger Numbers are quoted from~\cite{gao2018reconet}.}
	\label{tab:sota}
	\begin{tabular}{l|ccccc|c|c}
		\hline
		Models & \textit{Alley\_2} & \textit{Ambush\_5} & \textit{Bandage\_2} & \textit{Market\_6} & \textit{Temple\_2} & \textit{Sum} & \textit{FPS} \\ \hline\hline
		Teacher~\cite{ruder2018artistic} & 0.0560 & 0.0751 & 0.0489 & 0.0956 & 0.0679 & 0.3435 & 4.67\\\hline
		Student~\cite{gao2018reconet} &&&&&&& \\
		- From scratch & 0.0746 & 0.0887 & 0.0575 & 0.0997 & 0.0815 & 0.4019 & \bf 183 \\ 
		- Ours &\textbf{0.0524} & \textbf{0.0676} & \textbf{0.0445} & \textbf{0.0779} & \textbf{0.0627} & \textbf{0.3050}  & \bf 183 \\ 
		\hline
		Student~\cite{gao2018reconet} w/ $\mathcal{L}_{temp}$ &&&&&&& \\
		- From scratch & 0.0701 & 0.0844 & 0.0535 & 0.0948 & 0.0758 & 0.3787 & \bf 183\\ 
		- Ours &\bf \textbf{0.0506} & \textbf{0.0643} & \textbf{0.0423} & \textbf{0.0770} & \textbf{0.0596} & \textbf{0.2938} & \bf 183 \\
		\hline\hline
		Chen~\textit{et al.}~\cite{chen2017coherent}\textsuperscript\textdagger & 0.0934 & 0.1352 & 0.0715 & 0.103 & 0.1094 & 0.5125 & 17.5\\
		Ruder~\textit{et al.}~\cite{ruder2016artistic}\textsuperscript\textdagger & 0.0252 & 0.0512 & 0.0195 & 0.0407 & 0.0361 & 0.1727 & 0.62 \\\hline
	\end{tabular}
\end{table}

\begin{table*}[tb]
	\centering
	\caption{Temporal error $e_{stab}$ with style \textit{Candy} for networks with different distillation losses.}
	\label{tab:abla_distill_loss}
	\begin{tabular}{l|cccccc}
		\hline
		Models & \textit{Alley\_2} & \textit{Ambush\_5} & \textit{Bandage\_2} & \textit{Market\_6} & \textit{Temple\_2} & \textit{Sum} \\ \hline
		Student & 0.0746 & 0.0887 & 0.0575 & 0.0997 & 0.0815 & 0.4019\\ 
		+ $\mathcal{L}_{res}$ & 0.0606 & 0.0764 & 0.0493 & 0.0870 & 0.0691 & 0.3424 \\
		+ $\mathcal{L}_{rank}^{input}$ & 0.0716 & 0.0862 &	0.0554 & 0.0950 & 0.0773 & 0.3855
		\\
		+ $\mathcal{L}_{res}$ + $\mathcal{L}_{rank}^{input}$ &\textbf{0.0524} & \textbf{0.0676} & \textbf{0.0445} & \textbf{0.0779} & \textbf{0.0627} & \textbf{0.3050} \\ 
		+ $\mathcal{L}_{res}$ + $\mathcal{L}_{rank}^{T}$ & 0.0601 & 0.0755 & 0.0488 & 0.0882 & 0.0689 & 0.3415 \\ 
		\hline
		Student w/ $\mathcal{L}_{temp}$ & 0.0701 & 0.0844 & 0.0535 & 0.0948 & 0.0758 & 0.3787 \\ 
		+ $\mathcal{L}_{res}$ & 0.0574 & 0.0729 & 0.0483 & 0.0870 & 0.0661 & 0.3317 \\ 
		+ $\mathcal{L}_{res}$ + $\mathcal{L}_{rank}^{input}$ &\textbf{0.0506} & \textbf{0.0643} & \textbf{0.0423} & \textbf{0.0770} & \textbf{0.0596} & \textbf{0.2938} \\ \hline
	\end{tabular}
\end{table*}

\begin{figure}[tb]
	\centering
	\includegraphics[width = 0.7\linewidth]{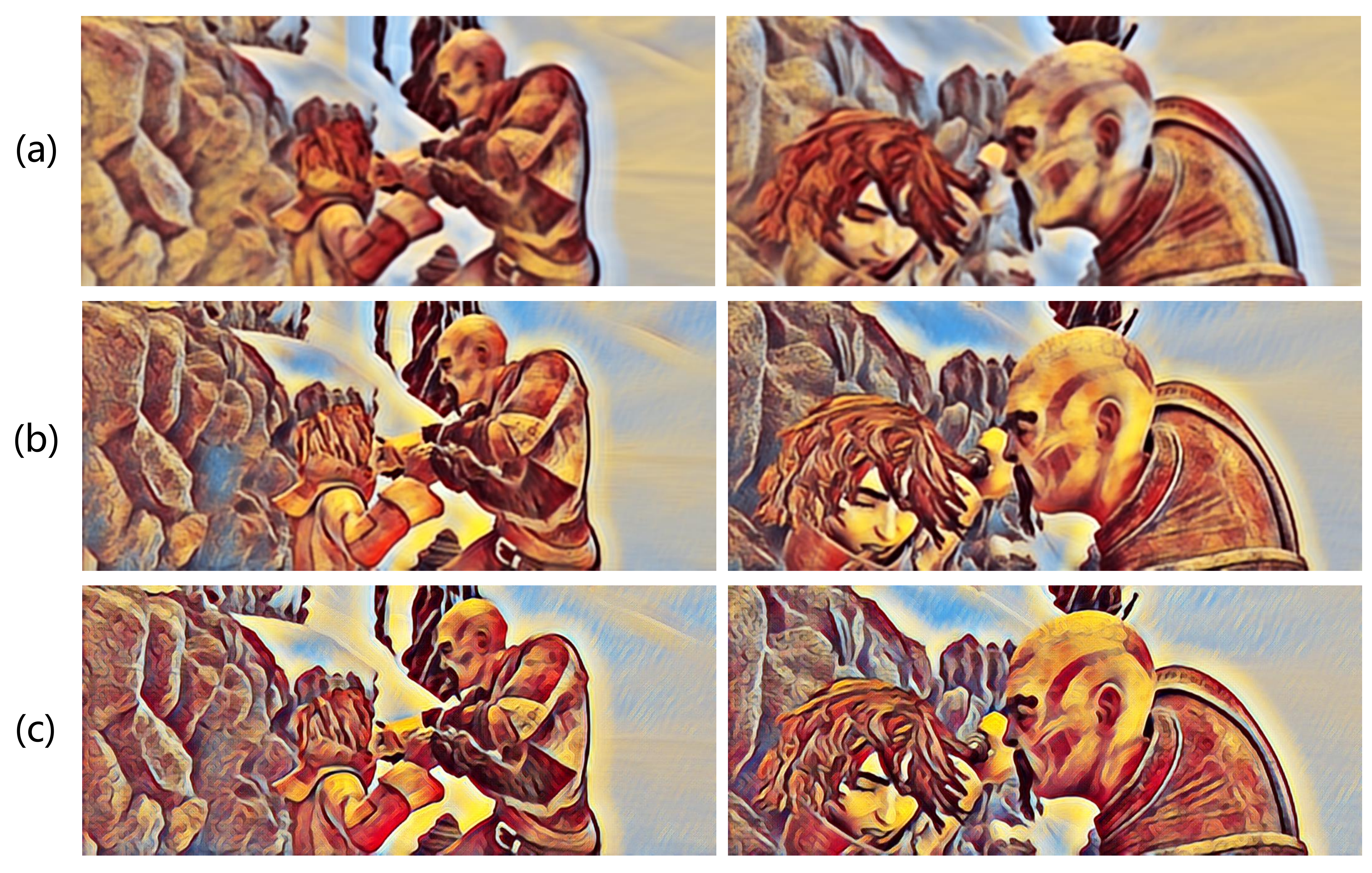}
	\caption{Training the student network with (a) vanilla distillation loss $\mathcal{L}_{vanilla}$, (b) residual distillation loss $\mathcal{L}_{res}$ and  (c) residual distillation loss and perceptual loss $\mathcal{L}_{percep} + \mathcal{L}_{res}$.}
	\label{fig:abla_res_kd}
\end{figure}
\begin{figure*}[tb]
	\centering
	\includegraphics[width = 1.0\linewidth]{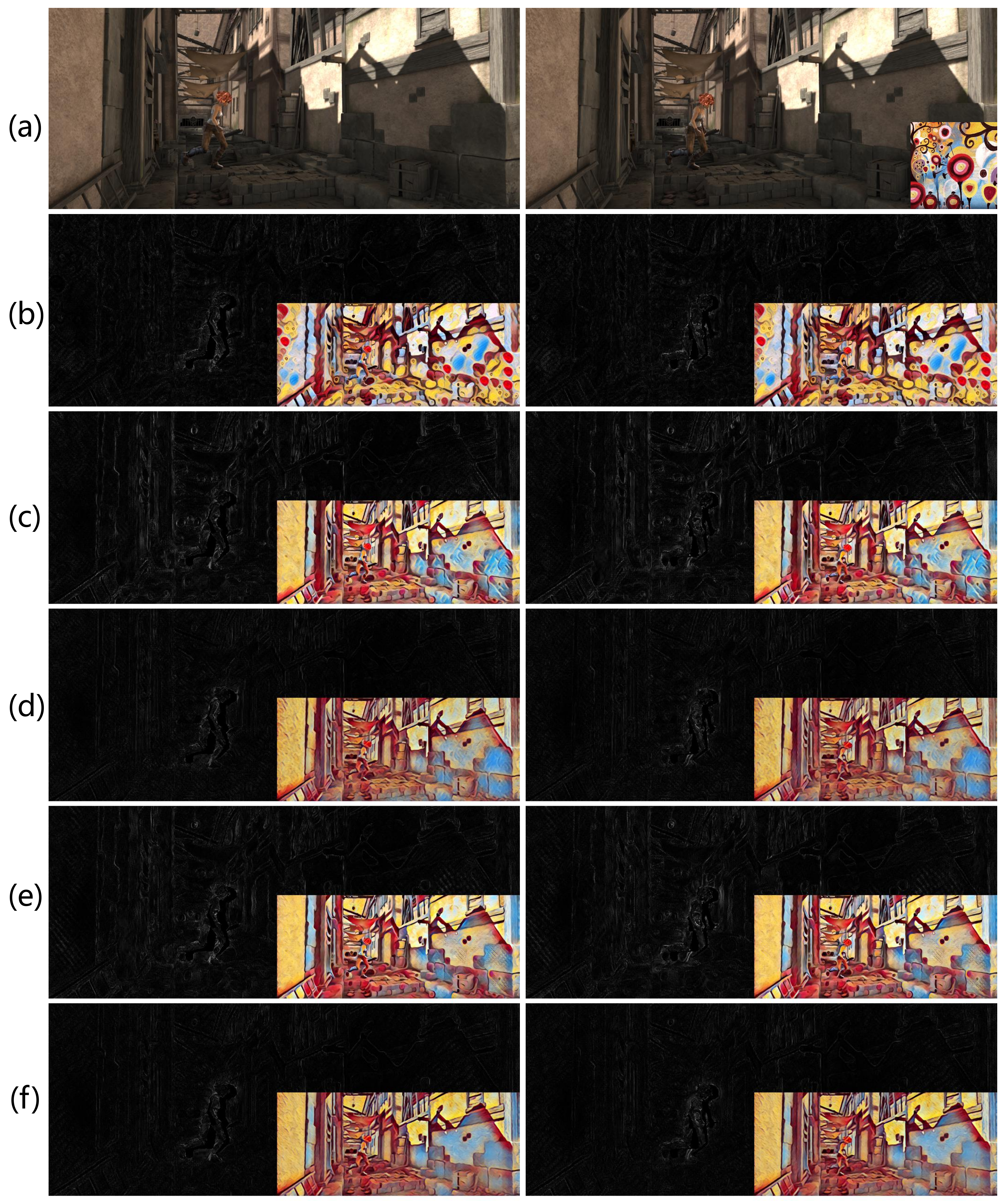}
	\caption{Qualitative results of different methods. (a) Two consecutive input video frames from MPI Sintel {Alley\_2} scene. The following rows show the temporal consistency errors of (b) Teacher network, (c) Student network trained from scratch, (d) Student network by our method, (e) Student network trained from scratch with additional $\mathcal{L}_{temp}$ and (f) Student network by our method with additional $\mathcal{L}_{temp}$. The temporal errors increase as shown from black to white in gray scale. Best viewed on screen.}
	\label{fig:qualitative}
\end{figure*}

\paragraph{\textbf{Quantitative Analysis.}} We compare our proposed methods with other video style transfer networks with style \textit{Candy} on five scenes from MPI Sintel Dataset. The temporal error $e_{stab}$ is calculated as Eq.~(\ref{eq:estab}) and is used to indicate the temporal consistency of output stylized videos. Output videos with smaller values of $e_{stab}$ are more stable. All inference speeds are evaluated on NVIDIA Tesla P100 GPU with the input size of $640 \times 320$. 

As shown in Table~\ref{tab:sota}, our proposed method consistently outperforms student baselines that are trained from scratch. For example, when we choose student network with only perceptual loss as baseline, training it with our proposed distillation losses improves $e_{stab}$ from $0.4019$ to $0.3050$. To further investigate the effectiveness of our proposed method, we then switch to a strong baseline, \ie,  using perceptual loss and temporal loss to train the student network. Training it from scratch using additional temporal loss gets more stable results. Nevertheless, our proposed method outperforms the vanilla student by a 22.4\% improvement for $e_{stab}$. When compared with the teacher network~\cite{ruder2018artistic}, our method achieves similar or better temporal consistency and runs much faster, since our method gets rid of time-consuming optical flow calculation during inference stage.

We further compare our proposed methods with state-of-the-art video style transfer methods~\cite{chen2017coherent,ruder2016artistic}. Table~\ref{tab:sota} shows that our method outperforms Chen~\textit{et al.}~\cite{chen2017coherent} for both temporal error $e_{stab}$ and running speed. Ruder~\textit{et al.}~\cite{ruder2016artistic} obtained smaller temporal error than our methods. However, it runs orders of magnitudes slower than our network.

\paragraph{\textbf{Qualitative Analysis.}} Qualitative results of different methods are shown in Fig.~\ref{fig:qualitative}. The first row shows two consecutive frames from the Alley\_2 scene of MPI Sintel Dataset. In this video, the viewpoint of the camera is continually changing. Meanwhile, only the person in the video moves and background regions remain unchanged. Fig.~\ref{fig:qualitative}~(b) shows the temporal consistency error of the baseline student network, \ie,  which is trained from scratch with perceptual loss. It shows that the student baseline produces less temporally consistent stylized frames. The results of our method are shown in Fig.~\ref{fig:qualitative}~(c)  and achieve better temporal consistency. A stronger student baseline, \ie,  trained with additional temporal loss, performs slightly better than the counterpart without temporal loss. Nevertheless, our method still obtains higher temporal consistency, which demonstrates the effectiveness of our proposed method. 

\subsection{Ablation Studies}

\paragraph{\textbf{Impacts of Distillation Loss.}} We first examine the impact of our proposed distillation losses, \ie,  residual distillation loss $\mathcal{L}_{res}$ and low rank loss $\mathcal{L}_{rank}^{input}$. As shown in Table~\ref{tab:abla_distill_loss}, residual distillation loss reduces $e_{stab}$ by 15\%. Adding low-rank distillation loss to the baseline network improve $e_{stab}$ from 0.4019 to 0.3855. Furthermore, utilizing low rank loss along with residual distillation loss reduces $e_{stab}$ by 24\%. For a stronger baseline student network that adopts temporal loss, residual distillation loss and low rank loss harvest consistent improvements. These experimental results demonstrate that our proposed distillation losses effectively improve the stability of output stylized videos.

\paragraph{\textbf{Discussion of Low-rank Loss.}} As we discuss in Section~\ref{subsec:lowrank}, there are two different design choices for low-rank distillation loss, \ie,  distilling low-rank knowledge from teacher network ($\mathcal{L}_{rank}^{T}$) and from input videos ($\mathcal{L}_{rank}^{input}$). As shown in the fifth and sixth row in Table~\ref{tab:abla_distill_loss}, learning low-rank information from input videos significantly outperforms the counterpart of learning from teacher network. It's not surprising since the stability of input videos is better than output videos of teacher network. Therefore, distilling low-rank information from input videos helps to improve the temporal consistency of student network.

\paragraph{\textbf{Residual KD vs. Vanilla KD.}} As we have discussed in Section~\ref{subsec:res_kd}, it's difficult to learn directly from the output of teacher network and thus produces blurry stylized images. As shown in Fig.~\ref{fig:abla_res_kd}~(a) and (b), and also in Fig.~\ref{fig:res_kd}, training student network with $\mathcal{L}_{res}$ in Eq.~(\ref{eq:res}) significantly improves the quality of stylized images compared with $\mathcal{L}_{vanilla}$ in Eq.~(\ref{eq:mse}).

\paragraph{\textbf{The Impact of Perceptual Loss.}} An intuitive question is that, can we simply train the student network with only the information of teacher network? Fig.~\ref{fig:abla_res_kd} shows that even if residual distillation loss alleviates the blurring problem, it is still not satisfying without the perceptual loss. Combining residual distillation loss and perceptual loss obtains better results with sharp edges and good style patterns. It is a reasonable observation since the task of style transfer has no groundtruth labels and it is difficult to force the student network to produce aligned outputs with teacher network in pixel-level. Therefore, the perceptual loss is still critical for the task of video style transfer knowledge distillation. 

\paragraph{\textbf{Impact of $K$.}} Here we discuss the impact of hyper parameter $K$, \ie,  the number of frames to calculate low rank loss. As shown in Table~\ref{tab:abla_k}, increasing $K$ from 3 to 4 slightly improves the temporal consistency of the output stylized videos. Further increasing $K$ to 5 obtains nearly saturated improvement. To balance the performance and training cost, we choose $K=5$ in our experiments.

\begin{table}[tb]
	\centering
	\caption{Temporal error $e_{stab}$ with style \textit{Candy} for different $K$ frames to calculate low rank loss.}
	\label{tab:abla_k}
	\begin{tabular}{l|ccc}
		\hline
		\qquad \qquad K & \qquad 3 & \qquad  4  & \qquad 5 \qquad\\ \hline
		Student (Ours) & \qquad 0.3158 &  \qquad 0.3061 & \qquad 0.3050\\\hline
		Student w/ $\mathcal{L}_{temp}$ (Ours) & \qquad 0.2986 & \qquad 0.2950 &\qquad 0.2938\\\hline
	\end{tabular}
\end{table}

\section{Conclusion}
\label{sec:con}

In this paper, we propose a novel method to distill knowledge from a stable video style transfer network with optical flow to a lightweight network that does not require optical flow during inference. In particular, we propose the residual loss to encourage student network to learn knowledge of the difference between the teacher networks with and without optical flow. Additionally, the low rank distillation loss is exploited to constrain the output stylized videos of student network to mimic the low rank of input videos, thus to further improve the stability.
Extensive experiments demonstrate that our proposed method achieves pleasing and stable stylized videos in high inference speed.

\paragraph{\textbf{Acknowledgments.}} We thank anonymous reviewers for their helpful comments. Chang Xu was supported by the Australian Research Council under Project DE180101438.

\clearpage
%
%
\bibliographystyle{splncs04}
\bibliography{egbib}
\end{document}